\documentclass[sigconf]{acmart}

\usepackage{booktabs} 
\usepackage{flafter}
\setcopyright{rightsretained}

\acmISBN{978-1-4503-9823-7}

\acmConference[ICVGIP'22]{13th Indian Conference on Computer Vision, Graphics and Image Processing}{December 2022}{Gandhinagar, India}
\acmYear{2022}
\copyrightyear{2022}

\acmPrice{15.00}

\editor{Soma Biswas}
\editor{Shanmuganathan Raman}
\editor{Amit K Roy-Chowdhury}

\acmArticle{83}

\copyrightyear{2022}
\acmYear{2022}
\setcopyright{acmlicensed}\acmConference[ICVGIP'22]{Proceedings of the Thirteenth Indian Conference on Computer Vision, Graphics and Image Processing}{December 8--10, 2022}{Gandhinagar, India}
\acmBooktitle{Proceedings of the Thirteenth Indian Conference on Computer Vision, Graphics and Image Processing (ICVGIP'22), December 8--10, 2022, Gandhinagar, India}
\acmPrice{15.00}
\acmDOI{10.1145/3571600.3571626}
\acmISBN{978-1-4503-9822-0/22/12}

\begin{document}

\title{A Fine-Grained Vehicle Detection (FGVD) Dataset for Unconstrained Roads}
\titlenote{Produces the permission block, and
  copyright information}

  \author{Prafful Kumar Khoba}
  \orcid{0000-0002-3853-9322}
  \affiliation{%
    \country{IIT Delhi, India}
  }
  \email{qiz228274@iitd.ac.in}
  
  \author{Chirag Parikh}
  
  \orcid{0000-0003-0216-7966}
  \affiliation{%
    \country{IIIT Hyderabad, India}
  }
  \email{chirag.parikh@students.iiit.ac.in}

 \author{Rohit Saluja}
    \orcid{0000-0002-0773-3480}
  \affiliation{%
    \country{IIIT Hyderabad, India}
  }
  \email{rohit.saluja@research.iiit.ac.in}

\author{Ravi Kiran Sarvadevabhatla}
\orcid{0000-0003-4134-1154}
  \orcid{1234-5678-9012}
  \affiliation{%
    \country{IIIT Hyderabad, India}
  }
  \email{ravi.kiran@iiit.ac.in}

  \author{C.V. Jawahar}
  \orcid{0000-0001-6767-7057}
  \affiliation{%
    \country{IIIT Hyderabad, India}
  }
  \email{jawahar@iiit.ac.in}

\renewcommand{\shortauthors}{}

\begin{abstract}
The previous fine-grained datasets mainly focus on classification and are often captured in a controlled setup, with the camera focusing on the objects. We introduce the first Fine-Grained Vehicle Detection (FGVD) dataset in the wild, captured from a moving camera mounted on a car. It contains 5502 scene images with 210 unique fine-grained labels of multiple vehicle types organized in a three-level hierarchy. While previous classification datasets also include makes for different kinds of cars, the FGVD dataset introduces new class labels for categorizing two-wheelers, autorickshaws, and trucks. The FGVD dataset is challenging as it has vehicles in complex traffic scenarios with intra-class and inter-class variations in types, scale, pose, occlusion, and lighting conditions. The current object detectors like yolov5 and faster RCNN perform poorly on our dataset due to a lack of hierarchical modeling. Along with providing baseline results for existing object detectors on FGVD Dataset, we also present the results of a combination of an existing detector and the recent Hierarchical Residual Network (HRN) classifier for the FGVD task. Finally, we show that FGVD vehicle images are the most challenging to classify among the fine-grained datasets.~\href{https://github.com/iHubData-Mobility/public-FGVD}{\textcolor{blue}{[GitHub]}}
\end{abstract}
%



\maketitle

\begin{figure}[th]
\includegraphics[width=\linewidth]{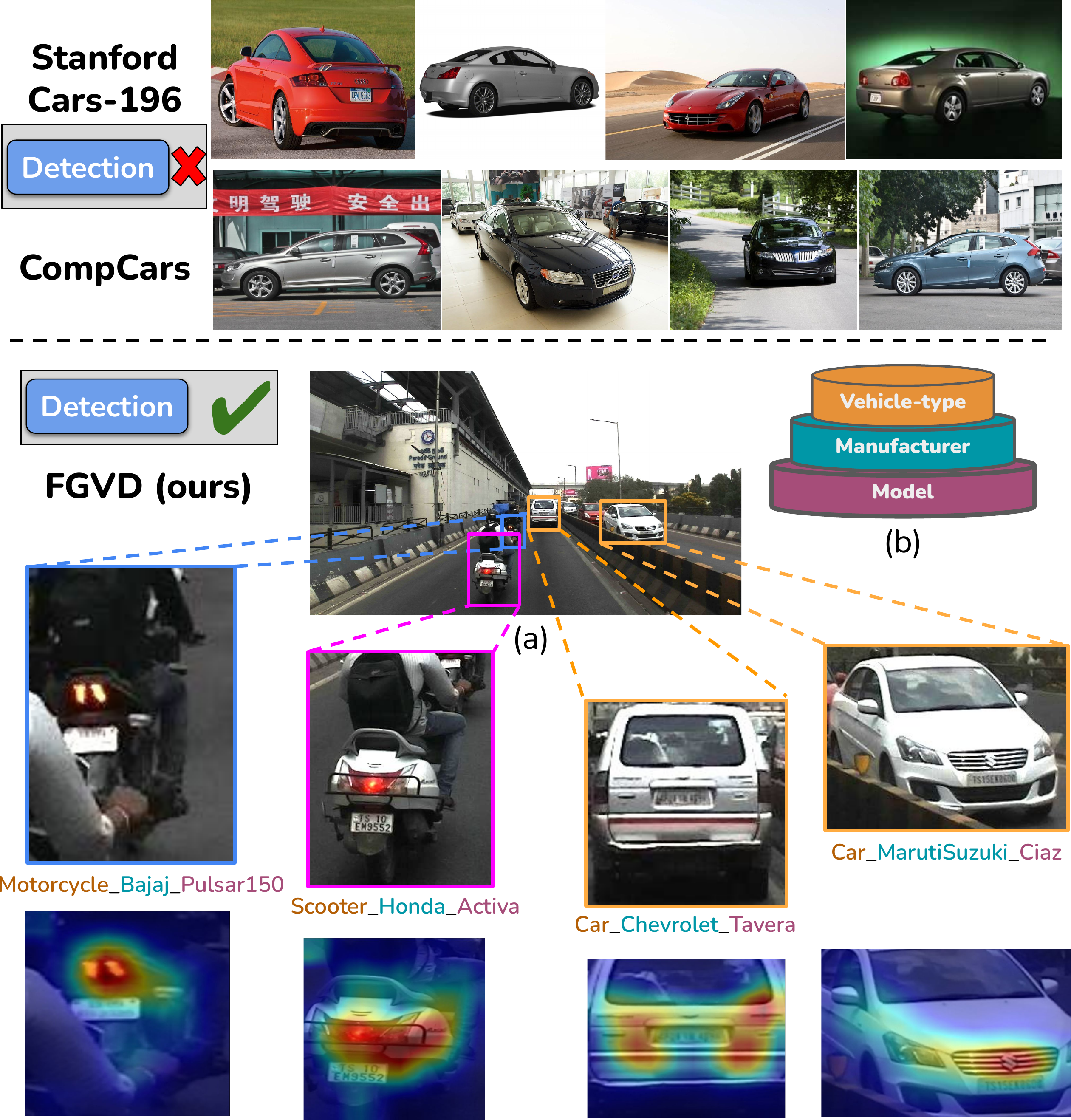}\Description{Diagram Figure}
\caption{Fine-grained datasets samples. Top: Previous datasets focus only on the classification of cars on vehicle-centric images. Middle: The proposed FGVD dataset enables fine-grained (multi) vehicle detection on unconstrained road scenes captured from vehicle-mounted cameras. Bottom left GradCAM++~\cite{DBLP:journals/corr/abs-1710-11063} visualizations on predicted crops show that the model focuses on the backlight and blinker at the motorcycle's top and scooter's bottom. For Tavera, the design on the left and right of license plates, and for Ciaz, the radiator and headlight regions are highlighted classification features.}
\label{fig:teaser}
\end{figure}\vspace{-5mm}
\begin{table*}
\begin{center}
\caption{Related Fine-grained datasets. All the previous road scene works include fine-grained classification only on car types. In contrast, FGVD has fine-grained detection labels for seven different types of vehicles (refer Fig.~\ref{fig:samples} and Table~\ref{table:heir}).}\vspace{-2mm}
\begin{tabular}{c c c c c c} 
{\bf Dataset } & {\bf Source} & {\bf \# Levels}  & {\bf \# Vehicle-type}& {\bf Classification} & {\bf Detection}  \\ \hline 
 BoxCars116k~\cite{Sochor2018} & CCTV & 3 & 1 & $\surd$ & $\times$  \\
 \hline
 CompCars~\cite{Yang_2015_CVPR} & Web/CCTV & 3 & 1 & $\surd$ & $\times$  \\
 \hline
 THS-10~\cite{najeeb2022fine} & CCTV & 2 & 1 & $\surd$ & $\times$ \\
\hline
Stanford Cars-196~\cite{KrauseStarkDengFei-Fei_3DRR2013} & Web & 2 & 1 & $\surd$ & $\times$ $^\psi$ \\
 \hline
{\bf FGVD (ours)} & {\bf Dashcam} & {\bf 3} & {\bf 6} & $\surd$ & {\bf $\surd$} \\
\hline
\multicolumn{6}{l}{$^\psi$ Stanford Car-196 is a part-based fine-grained car recognition dataset containing bounding boxes for cars' parts.}
\end{tabular}\vspace{-3mm}
\label{table:related_datasets}
\end{center}
\end{table*}
\section{Introduction}
\begin{figure*}[t]
\makebox[\textwidth]{\includegraphics[width=\textwidth]{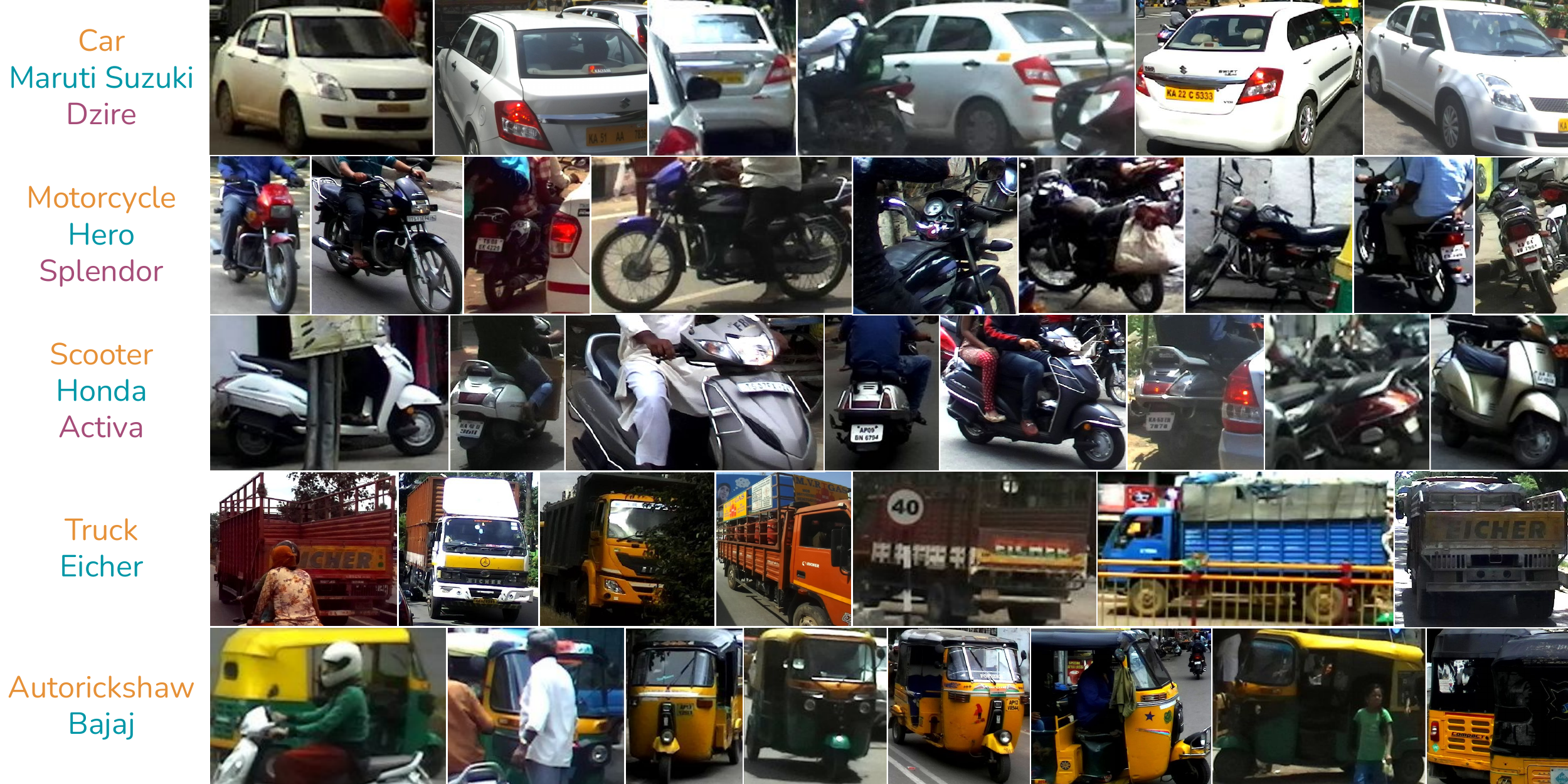}}\Description{Depiction}
\caption{Sample images of different categories in FGVD exhibiting inter-class similarities (bottom two rows), multiple vehicle orientations (all rows), and frequent occlusions.}
\setlength{\belowcaptionskip}{-10pt}
\label{fig:samples}
\end{figure*}

Intelligent traffic monitoring systems are of utmost need in big cities for public security, planning, and surveillance. For the tasks like vehicle re-identification and robust detection (e.g., when a vehicle occludes another vehicle that is similar in appearance), the detectors used in the surveillance systems should finely classify the vehicle type, manufacturer, and model of the on-road vehicles. Conventionally, detection models like YOLO~\cite{yolov3} and Faster R-CNN~\cite{fasterRCNN} are trained to classify vehicles based on coarse categories of on-road datasets like BDD and IDD~\cite{yu2020bdd100k,varma2019idd}. A coarse class can contain multiple sub-classes with minute variations, referred to as the fine-grained classes. The localization of such sub-class vehicles based on their granularity in the design is known as Fine-Grained Vehicle Detection (FGVD). The FGVD models and datasets can enable robust vehicle re-identification and detection in highly dense and occluded traffic scenarios. Therefore, we propose a novel FGVD dataset with multiple hierarchy levels for the fine-grained labels. Fig.~\ref{fig:teaser} depicts a sample scene image from the FGVD dataset and the corresponding labels. As shown in the figure, in addition to enabling the detection task, the FGVD dataset includes complex intra-class and inter-class variations in types, scales, and orientations compared to the previous fine-grained classification datasets. The dataset also contains challenging occlusion scenarios and lighting conditions  (refer to Fig.~\ref{fig:samples}). FGVD comprises three levels of hierarchy, i.e., vehicle type, manufacturer, and model, as shown at the bottom of the vehicles' Regions of Interest (ROIs) in Fig.~\ref{fig:teaser} (a) and in Fig.~\ref{fig:teaser} (b) with three different colors.

While some classification datasets like Stanford Cars-196~\cite{KrauseStarkDengFei-Fei_3DRR2013} and CompCars~\cite{Yang_2015_CVPR} also includes the makes for different kinds of cars (see Fig.~\ref{fig:teaser} top), we introduce the complementary hierarchical labels for two-wheelers, autorickshaws, trucks, and buses (refer Sec.~\ref{section:Dataset Collection}). The granularity of the FGVD dataset increases as we move from parent to child level (refer to Figs.~\ref{fig:teaser} (b) and~\ref{fig:heirtree}). For the hierarchical FGVD dataset, every level has its uniqueness. Firstly, different vehicles may look similar at the first level of granularity, e.g., motorcycles and scooters, both being the two-wheelers. However, as it can be inferred from Fig.~\ref{fig:teaser}, the overall design of the scooter is different from the motorcycle, e.g., scooters have a backlight at the bottom as compared to the top backlight of the motorcycle. The overall appearance of the two vehicles with the same parent may look even more similar. However, the minute subtle and local differences are present in the same subcategory. Also, some categories are not present in the earlier vehicle datasets, for example, scooter, autorickshaw, truck, and bus; hence, they must be added to a fine-grained dataset. Therefore, we introduce classes unique to the FGVD dataset to facilitate detailed research in fine-grained on-road scenarios. The main contributions of this work are as follows:
\begin{itemize}
    \vspace{-0.2cm}\item A novel Fine-Grained Vehicle Detection (FGVD) dataset for on-road vehicles in dense and occluded traffic scenarios. To the best of our knowledge, no fine-grained detection dataset exists in the literature, and ours is the first dataset of its kind.
    \item We present the results of baseline detection and classification models on the proposed dataset. We also show the initial results of a combination of a detector and a recent hierarchical fine-grained classification model.
\end{itemize}
\vspace{-0.2cm}
\begin{figure*}
    \makebox[\textwidth]{\includegraphics[width=0.9\textwidth]{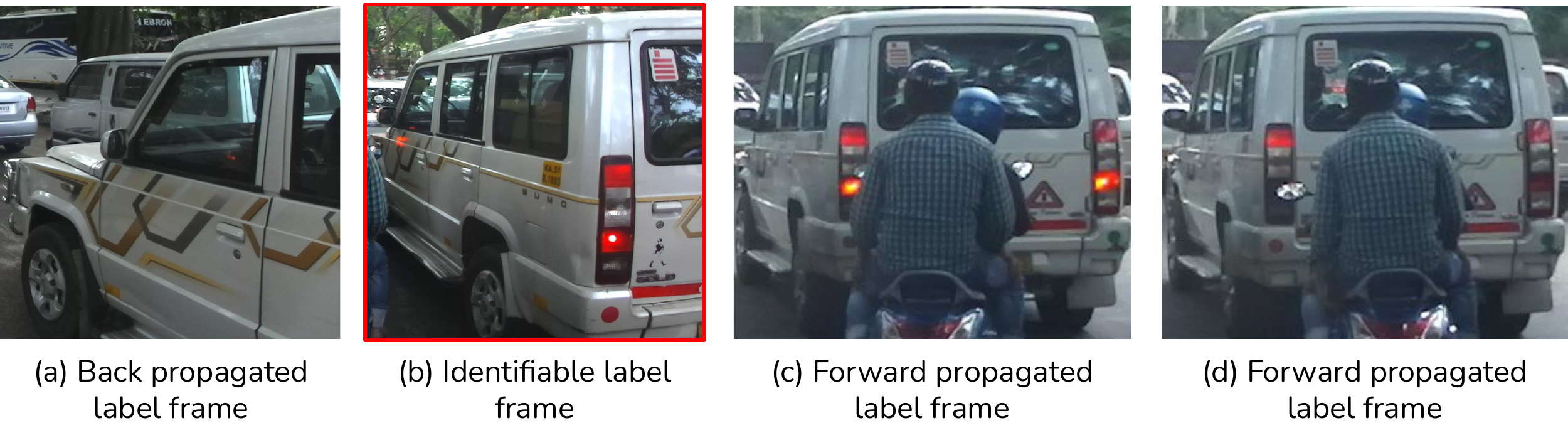}}
    \caption{Annotation Strategy: (b) is the only confidently identifiable image frame where the label cue (logo at bottom-left of the car) is visible. As the frames are from common video sequence, the label is propagated to the vehicle instances in (a), (c) \& (d).}
    \label{fig:anno}
\end{figure*}
\section{Related Work}
{\bf Fine-grained Classification Datasets:} Early fine-grained classification datasets mainly focus on birds~\cite{wah2011caltech}, flowers~\cite{nilsback2008automated}, dogs~\cite{KhoslaYaoJayadevaprakashFeiFei_FGVC2011, Horn_2015_CVPR}, aircrafts~\cite{maji2013fine}. The datasets like birds~\cite{wah2011caltech} cover appropriate occlusion examples, and the aircrafts dataset~\cite{maji2013fine} has objects at low resolution showcasing sufficient complexity for the classification task. But the objects in these datasets are usually located around the image center, making it unfit for their detection in the wild. There are few public datasets for fine-grained vehicle classification with hierarchical labels of the car’s make, model, submodel, and year of manufacture~\cite{Sochor2018, najeeb2022fine, Yang_2015_CVPR}. Jakub et al. ~\cite{Sochor2018} publish the BoxCars116k dataset with $3D$ bounding box annotations of $1,16,286$ car images with $693$ fine-grained classes. Yang et al.~\cite{Yang_2015_CVPR} released the CompCars dataset, which contains $44,481$ frontal view images of cars taken from surveillance cameras. It also contains web-nature images from different viewpoints of $136K$ vehicles classified into $600$ categories. Najeeb et al.~\cite{najeeb2022fine} released $4250$ CCTV car images of $10$ different models. Similar to this, the Stanford Cars dataset~\cite{KrauseStarkDengFei-Fei_3DRR2013} includes $16,185$ vehicle images of $197$ car types and their part labels to assist in fine-grained recognition tasks. All the above-mentioned recognition datasets contain only one car per image and lack the complexity required for detection in real-world traffic scenes. Our detection dataset includes diverse traffic scenarios observed in urban settings capturing large variations in scale, pose, occlusion, illumination, and density of vehicles. All the vehicle classification datasets mentioned above have fine-grained labels only for cars. In contrast, our dataset introduces these labels for four additional vehicles - motorcycles, scooters, autorickshaws, and trucks. Also, the proposed FGVD dataset contains images captured from dashboard cameras installed on top of surveillance vehicles instead of using static CCTV cameras, making its use economically viable and sustainable for road safety in any remote city location. The datasets like BDD100k~\cite{yu2020bdd100k}, Waymo ~\cite{Sun_2020_CVPR}, and IDD~\cite{varma2019idd} do focus on vehicle detection but lack fine-grained labels.

{\bf Detection and Fine-grained Classification Models:}
Recent detection models like Yolov5~\cite{yolov5} and Faster-RCNN~\cite{fasterRCNN} consider all object labels as independent from each other. So, they do not model the hierarchical relationship between the object’s fine-grained labels. The difficulty of detecting fine-grained objects thus increases with deeper class definition as the number of samples per class becomes smaller, and the visual cues become more challenging. Chen et al.~\cite{chen2022label} recently proposed Label Relation Graphs Enhanced Hierarchical Residual Network (HRN), which gives state-of-the-art performance on fine-grained classification datasets. Their architecture exploits the parent-child correlation between labels by transferring the hierarchical knowledge through residual connections across feature levels, but it is not used for object detection. Therefore, we combined the object detection and classification models and obtained superior performance on fine-grained vehicle detection.

\section{FGVD Dataset and Annotation}\label{section:Dataset Collection}
We use images with corresponding coarse labels, and bounding boxes from the IDD detection dataset  ~\cite{varma2019idd}. The vehicles far from the camera are infeasible to annotate. Hence, we remove the bounding boxes with a height-to-width ratio lesser than the thresholds, which are different for different vehicle types. To consider the variability of vehicles in physical dimensions, we keep the threshold for the truck's bounding box ratio higher than that of the car, which in turn is higher than that of the bike. The thresholding process makes the annotators' work easy, manageable, and quick. The FGVD dataset contains $5502$ scene images containing around 24450 bounding boxes, with $217$ ($210$ unique, and $7$ repeated from higher levels) fine-gained labels in the third level. 

\subsection{Annotation Process}\label{section:Annotations&Quality check}
We select $5502$ out of $16311$ high-quality images from the IDD-Detection dataset based on occlusion, size of vehicle boxes, and traffic density. The annotation team consists of four highly skilled annotators and two expert reviewers for quality checks. Firstly, we train the annotators for the task by providing fine-grained labels for a few FGVD  samples. Secondly, we provide the guidelines, template, and a list of objects to be annotated. After proper training, the annotators can recognize the popular vehicles in the scene. However, if still, the vehicle is not recognizable, then they can use google lens or image search on the internet. For example, consider the scenario where the annotator can recognize the manufacturer by looking at the brand logo. Still, the model name is not visible due to occlusion, truncation, or any other complexity. In such scenarios, the annotator would search for similar vehicles on the manufacturer's website. The essential part of looking at the image for classification is the vehicle's overall design, design of its components (e.g., some scooters have petrol openers at the back), brand logo, and model name. While creating IDD-Detection dataset~\cite{varma2019idd}, many images are taken from the continuous video frames; therefore, images have a temporal connection. For instance, in Fig.~\ref{fig:anno} Tata Sumo is not confidently identifiable in the first frame due to truncation, but in the second frame, the brand label is visible, which gives the confidence to the annotator to label even when they are unable to recognize the vehicle's design. Similarly, the annotators propagate the label in the next frame, in which there is a lot of occlusion from other vehicles, and it is comparatively hard to annotate if they do not connect the knowledge from different frames.

We divide the data creation process into two steps, i.e., the pilot phase and the takeoff phase. In the pilot phase, each annotator labels a small set of images containing less traffic density. We also train the annotators to label the vehicles from the dataset images, which are confidently identifiable but do not have a bounding box. In the case of cars, motorcycles, and scooters, we have an attribute called \emph{new}. Whenever annotators encounter a new variant of any model in our dataset, they tickmark the \emph{new} attribute in the checkbox. During the pilot phase, we note the average time to annotate one image, from which we estimate the number of days for creating the whole dataset. Performance in the pilot phase lets us choose the best two annotators as reviewers in the next stage. In the takeoff phase, many images contain high traffic density, which in turn causes massive occlusion (samples shown in Fig.~\ref{fig:samples}). All the labels in the takeoff stage are reviewed. The images having labels with high confidence from the reviewers are selected to be part of the dataset. The remaining vehicle's bounding boxes, for which any fine-grained levels are ambiguous, are labeled as ``others". If any material like a vehicle cover or a cloth covers any vehicle, they mark it as ``covered". The average time to annotate one image is one minute and half a minute to perform a quality check. We annotate all the scenes in the FGVD dataset using the Computer Vision Annotation Tool (CVAT)\footnote{
\href{https://github.com/openvinotoolkit/cvat}{\textcolor{blue}{https://github.com/openvinotoolkit/cvat}}}
\\
\begin{figure}[ht]
\includegraphics[width=\linewidth]{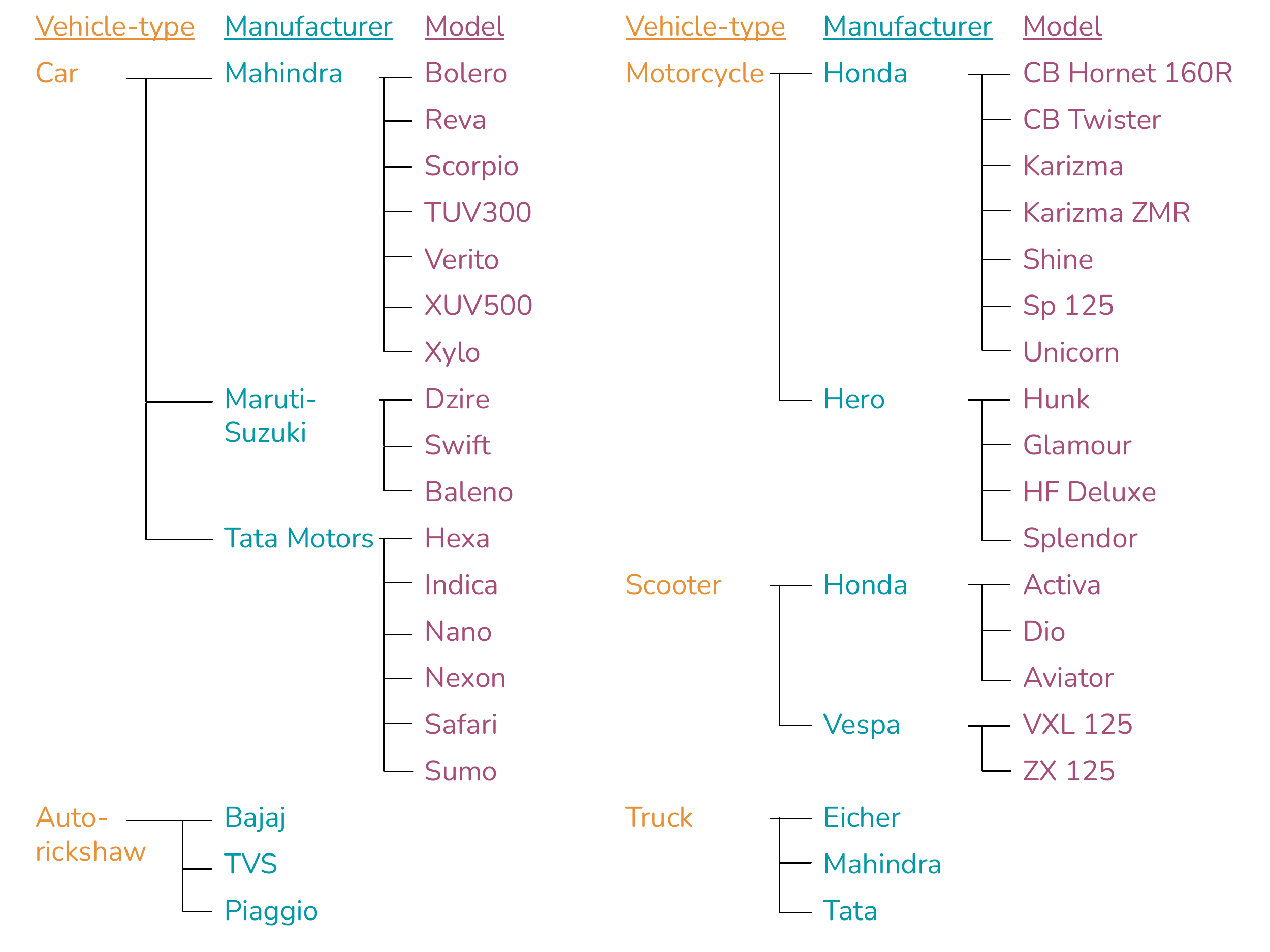}\Description{Moneyshot Figure}
\caption{Sample Hierarchy Tree of the FGVD dataset.}
\label{fig:heirtree}\vspace{-4mm}
\end{figure}

\begin{table}
    \begin{center}
    \begin{tabular}{cccc}
    \toprule
    Vehicle Type & Levels of Hierarchy & L-2 labels & L-3 labels \\\midrule
    Car          & 3    & 22 &  112               \\\hline
    Motorcycle   & 3    & 11 &  67               \\\hline
    Scooter      & 3    & 9 &   23               \\\hline
    Truck        & 2    & 7 &   7               \\\hline
    Autorickshaw & 2    & 6 &   6               \\\hline
    Bus          & 2    & 2 &   2 \\\hline             
    {\bf Total}  & {\bf 3}    & {\bf 57} &   {\bf 217} \\             
    \bottomrule
    \end{tabular}
    \end{center}
    
    \caption{Levels of Hierarchy for different Vehicles in FGVD.}
    
    \label{table:heir}\vspace{-4mm}
\end{table}

\begin{figure}
\includegraphics[width=0.8\linewidth]{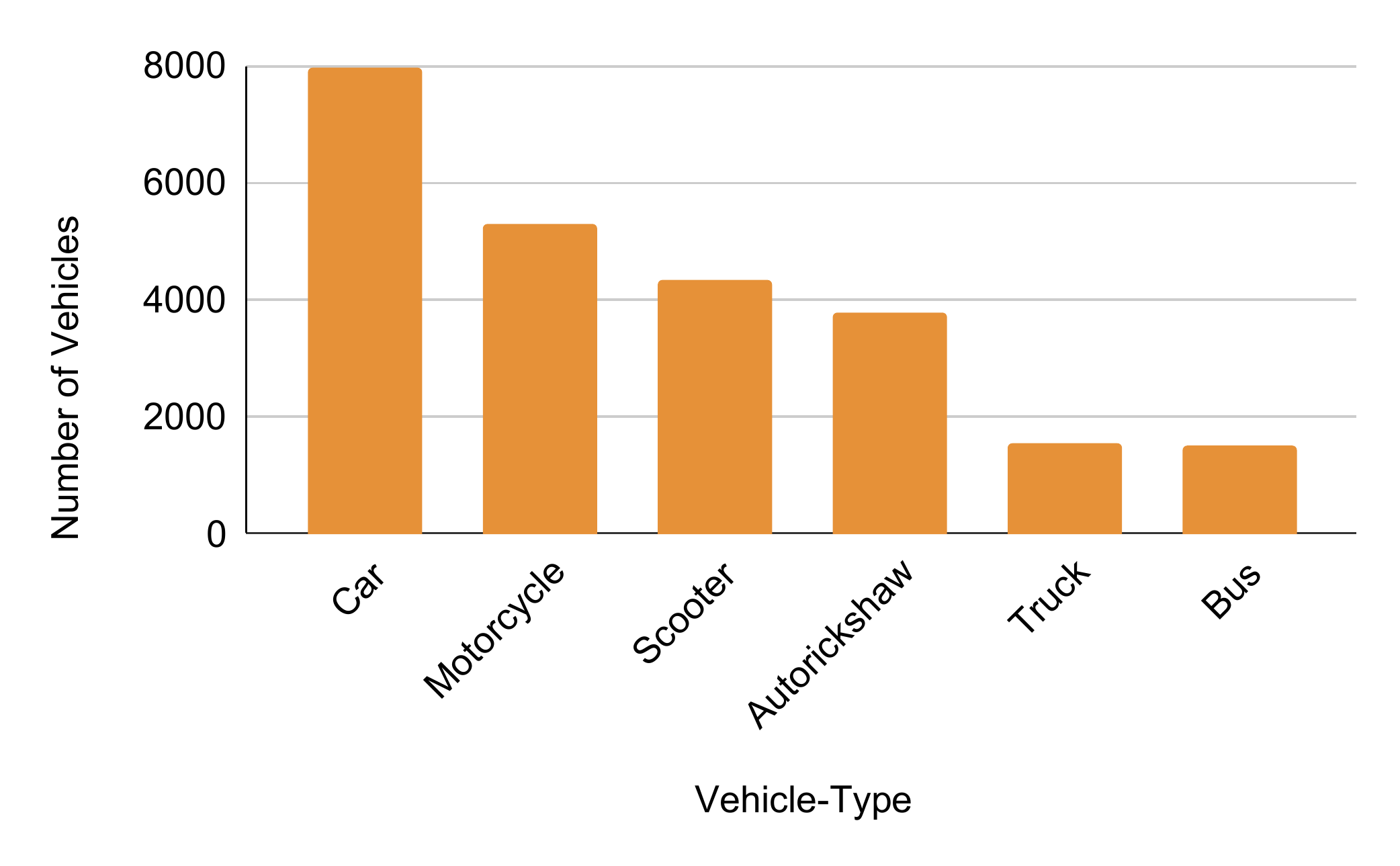}\Description{Moneyshot Figure}\vspace{0.1mm}
\includegraphics[width=0.8\linewidth]{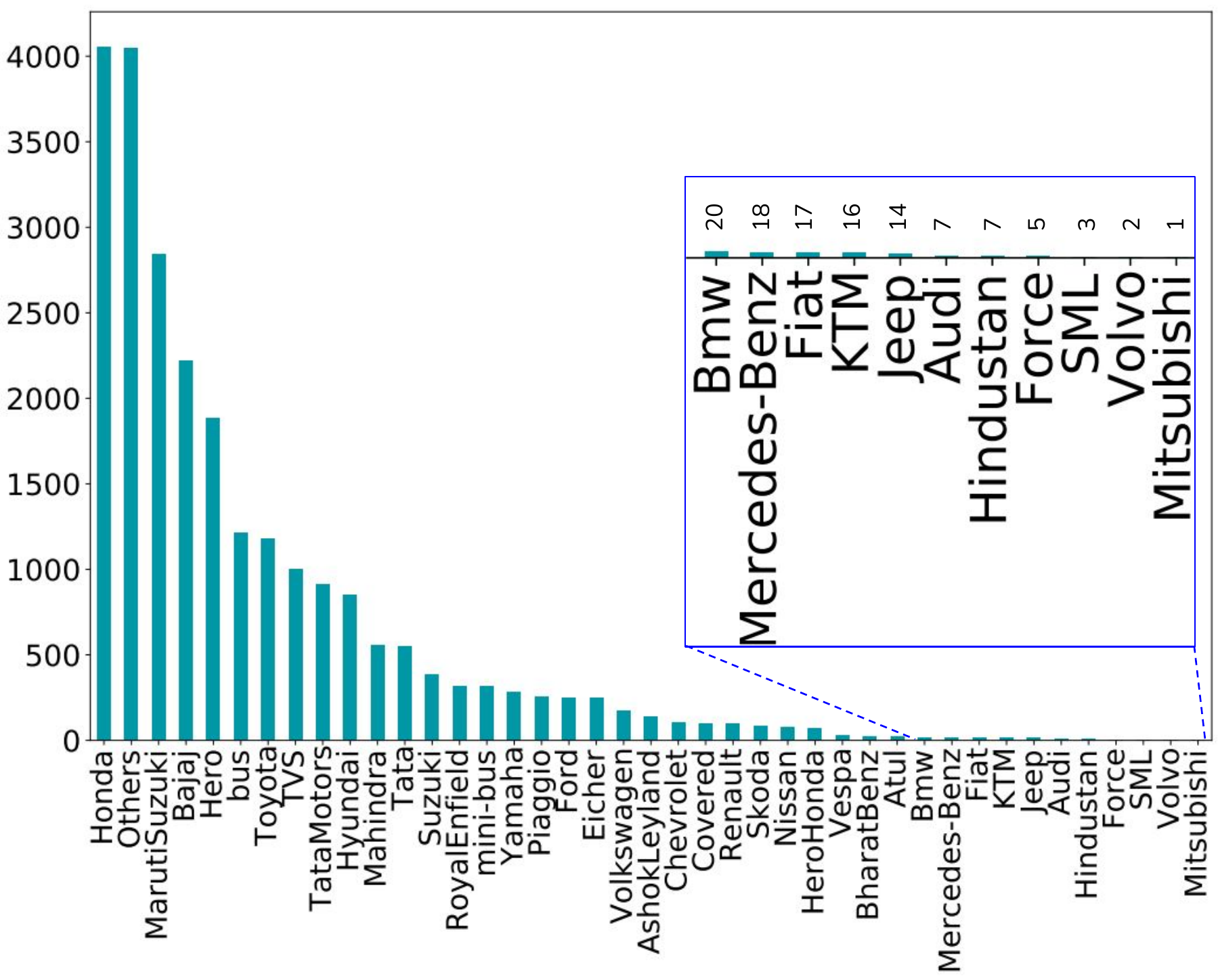}
\vspace{-4mm}
\caption{Histograms for Level-1 (top) and Level-2 (bottom) labels in FGVD.}
\setlength{\belowcaptionskip}{-10pt}
\label{fig:histograms}\vspace{-4mm}
\end{figure}

\begin{figure*}[t]
\makebox[\textwidth]{\includegraphics[width=0.9\textwidth]{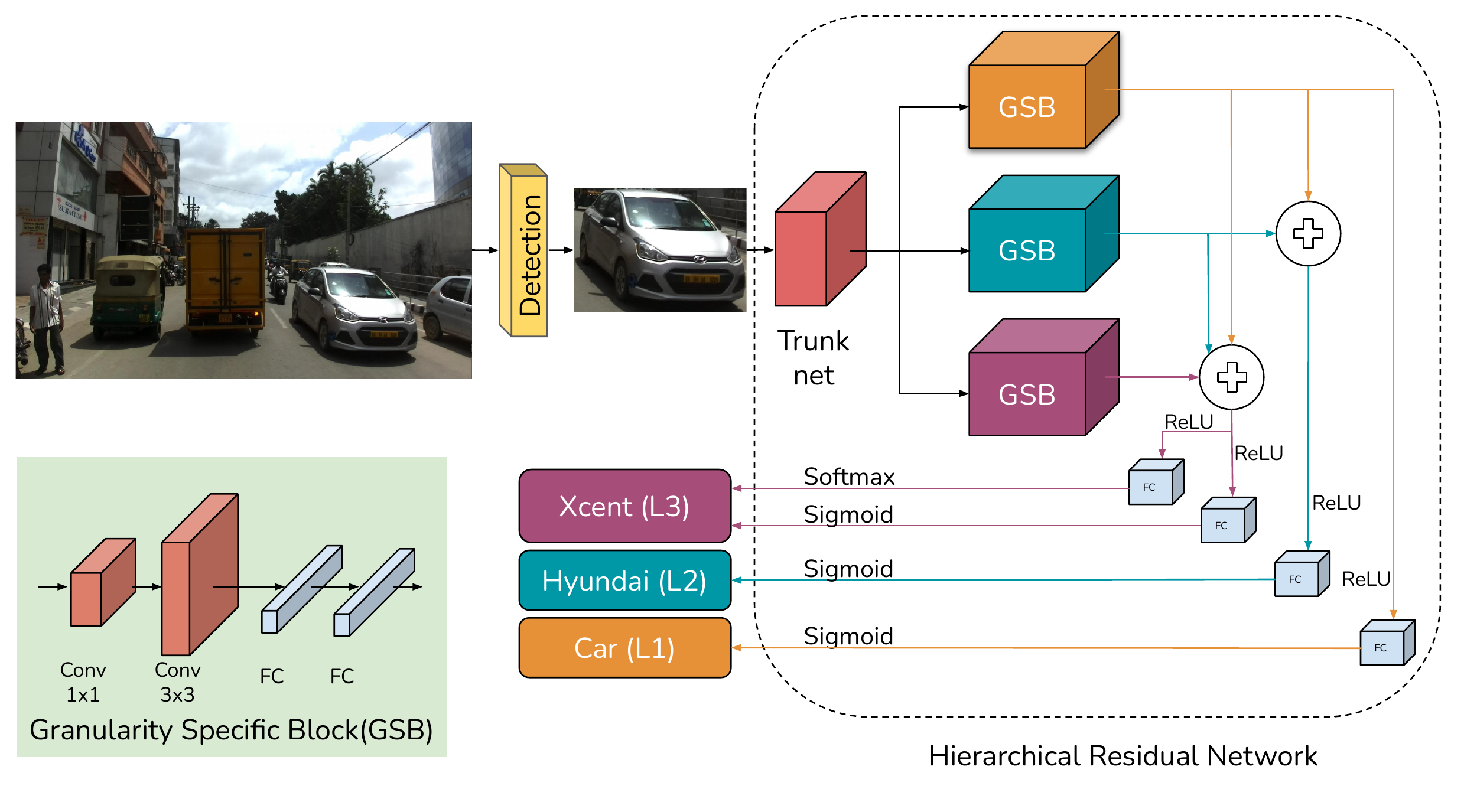}}\Description{Depiction}\vspace{-5mm}
\caption{Architecture for fine-grained vehicle detection. We use YOLO~\cite{yolov5} for localization and Label Relation Graphs Enhanced HRN~\cite{chen2022label} for classification.}
\label{fig:architecture}
\end{figure*}
\subsection{Hierarchy and Long-tailed Distribution}\label{section:Heirarchy and Long-tailed distribution}
As shown in Table~\ref{table:heir}, each vehicle type in the proposed Fine-Grained Vehicle Detection (FGVD) dataset has different hierarchical levels. 
We illustrate the sample vehicle images and hierarchy tree of the FGVD dataset in Figs.~\ref{fig:samples} and~\ref{fig:heirtree}. As shown, the FGVD contains three different levels of hierarchy, which we detail below\footnote{For bus class, we do not follow this hierarchy. Small buses, known as ``mini-bus," and the ``general-bus'' are the child classes for the bus.}:
\begin{itemize}

    \item {\bf Vehicle-type:} The highest coarse level labels of the vehicle come under the vehicle-type category. We consider it as level 1 of the hierarchy. Car, motorcycle, scooter, truck, auto-rickshaw, and bus are the six categories present in vehicle type.
    \item {\bf Manufacturer:} The manufacturer level contains the primary producer of the vehicles. The manufacturer category has finer details than the vehicle type level. A producer may manufacture multiple kinds of vehicles. For example, Bajaj manufactures motorcycles as well as auto-rickshaw.
    \item {\bf Model:} The model level is at the last group of the hierarchy. This level comprises highly fine-grained features that are unique for the variant. For example, a car's design must be unique for each manufacturer.
\end{itemize}
As illustrated in Fig.~\ref{fig:histograms}, 
the annotation levels contain the common challenge of a class imbalance to different degrees. %

\section{Methodology}\label{section:Methodology}
Fine-Grained Vehicle Detection (FGVD) aims to localize the vehicles in an on-road scene image and identify their type, manufacturer, and model variant. We accomplish this in two stages: the first stage involves vehicle localization, and the second involves fine-grained classification of the localized object. The entire pipeline is shown in Fig.~\ref{fig:architecture}. In the localization stage, we use YOLOv5~\cite{yolov5} model, which gives us the vehicle bounding boxes. We then crop out the vehicles' Regions Of Interest (ROIs) from the original image using the bounding boxes obtained in the localization stage. The cropped ROIs are then resized before feeding them to the classification module. In the classification module, we use the Label Relation Graphs Enhanced Hierarchical Residual Network (HRN) model~\cite{chen2022label}, which predicts the coarse to fine-grained classes for the ROIs.

\subsection{Vehicle Localization}
We train the YOLOv5 model to localize the vehicles in the FGVD dataset\footnote{For training the HRN model, we prepare the dataset separately by cropping out vehicle images using the ground truth boxes from our dataset.}. There are various reasons for choosing the YOLOv5 model for vehicle localization. Firstly, YOLOv5 incorporates Cross Stage Partial Network (CSPNet) ~\cite{DBLP:journals/corr/abs-1911-11929} into its backbone and in the neck. The CSPNet helps to achieve a richer gradient combination while reducing the amount of computation, which ensures the inference speed and accuracy are high and reduce the model size.

Moreover, the HRN's classification accuracy depends on the localization model's performance; thus, maintaining high accuracy for vehicle localization is essential. Secondly, the head of YOLOv5 generates three different sizes (18×18, 36×36, 72×72) of feature maps to achieve multi-scale~\cite{DBLP:journals/corr/abs-1804-02767} prediction, enabling the model to handle small, medium, and large-sized objects. 
YOLOv5 also auto-learns custom anchor boxes such that the anchors are adapted to our FGVD dataset, which helps improve the detection results. It also incorporates various augmentations, such as mosaic, during training which significantly helps to generalize. Moreover, we experiment with Faster-RCNN~\cite{fasterRCNN}, but we obtain the best results with YOLOv5 (refer to Sec.~\ref{section:RESULTS}) while it also takes the least training time.

\subsection{Fine-grained Classification}
For several reasons, we chose the Label Relation Graphs Enhanced HRN network~\cite{chen2022label} for fine-grained vehicle classification. The HRN network focuses on encoding the label hierarchy from coarse-to-fine levels. The HRN accomplishes this by using the Granularity Specific Blocks (GSB) and residual connections, as shown in Fig 4. Each GSB block extracts the hierarchical level features by processing the feature maps generated from the trunk network, i.e., any common feature extraction network pre-trained on ImageNet~\cite{5206848}. The residual connections combine the features of coarse-level and fine-level subclasses. This kind of hierarchical modeling primarily benefits the FGVD application on our dataset because there are many similarities between different vehicle model variants corresponding to the same manufacturer or vehicle type. The HRN model incorporates a combinatorial loss which aggregates information from related labels defined in the tree hierarchy. This tree hierarchy uses a sigmoid node for each label, which can be seen in Fig.~\ref{fig:architecture} for the L-1, L-2, and L-3 outputs. The HRN models independent relations using sigmoid instead of softmax since sigmoid implies mutual exclusion. But if the training samples at the fine-grained levels are few, the combinatorial loss would fail to well-separate the skewed classes. So, an additional multi-class cross-entropy loss is used with the softmax function for the finest labels, depicted for L-3 class output in Fig.~\ref{fig:architecture}. The softmax function increases the weightage of fine-grained classification loss, ultimately ensuring high classification accuracy specifically for the fine-grained labels. 
Moreover, the residual connections in the HRN for hierarchical feature interactions make the architecture effective compared to other models while demonstrating state-of-the-art performance on standard fine-grained classification datasets. 

We also create a hierarchical tree structure of labels for each of our fine-grained classes in the format required by the HRN architecture. It is important to note that we use the softmax output for the fine-grained class instead of the sigmoid output in the HRN model at inference time. As mentioned by Chen et al.~\cite{chen2022label}, the softmax output channel computes separate cross-entropy loss so that the mutually exclusive fine-grained classes gain more attention during training.
\begin{figure*}[ht]

\includegraphics[width=0.45\linewidth,height=6cm]{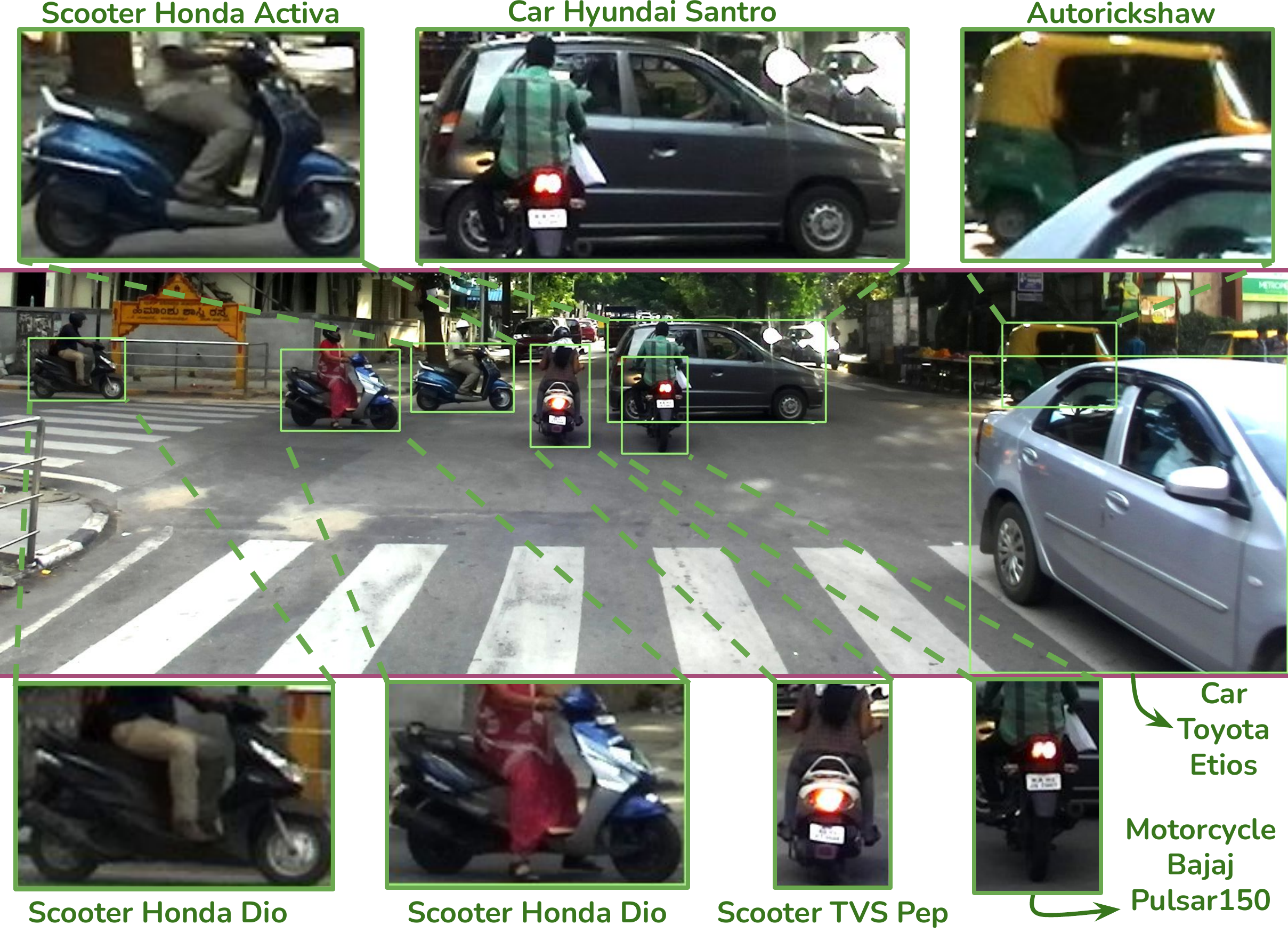}\Description{Diagram Figure}
\hspace{5mm}
\includegraphics[width=0.5\linewidth,height=6cm]{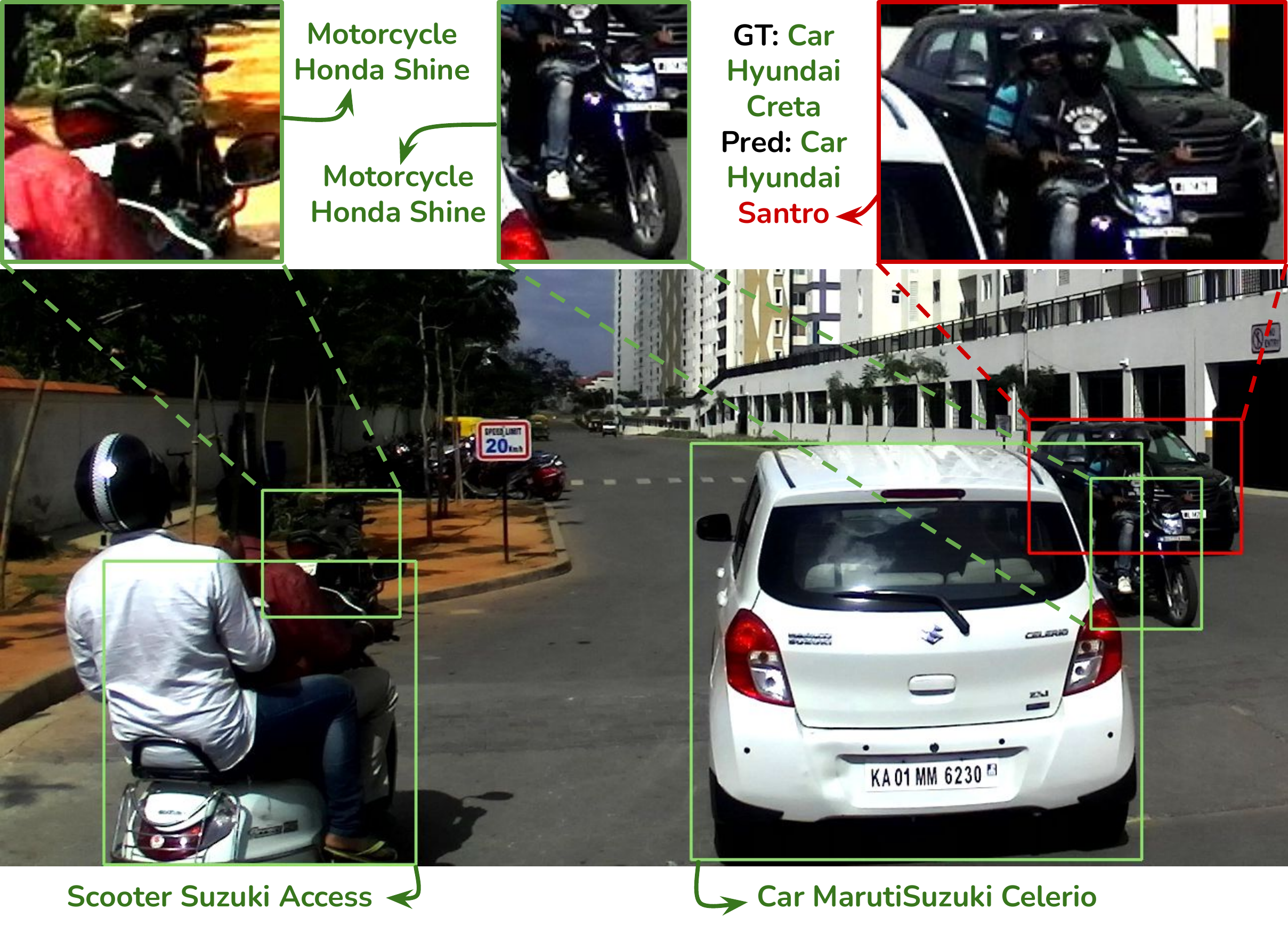}\Description{Diagram Figure}

\vspace{-2.5mm}
\setlength{\belowcaptionskip}{-10pt}
\caption{FGVD results of the YOLOv5l+HRN model on sample scene (crops shown for better visibility) images from the test dataset. The detected bounding boxes are marked in green and red, corresponding to correct and incorrect class predictions. The zoomed-in views highlight the vehicles which are occluded or have low visibility.}
\label{fig:fgvd_sample_outputs}
\end{figure*}
\section{EXPERIMENTS}\label{section:EXPERIMENTS}
We split our entire dataset into train:val:test ratio of 64:16:20. We use the YOLOv5l~\cite{yolov5} model pre-trained on the COCO dataset~\cite{DBLP:journals/corr/LinMBHPRDZ14} and fine-tune it on the FGVD dataset for $100$ epochs with a batch size of $8$. The input images are pre-processed and resized to $640\times640$ pixel dimensions before feeding them to the training pipeline. While training, we observe that the objectness loss on the validation set reaches the lowest point after a few epochs and then starts diverging. To resolve this, we reduce the contribution of the objectness loss to the overall loss function by half. 
We train the HRN model on the ground truth Regions of Interest (ROIs) and all three FGVD levels. 
We use the resenet-50~\cite{DBLP:journals/corr/HeZRS15} model, pre-trained on ImageNet~\cite{5206848}, for the trunk net in the HRN architecture. The input image size used here is 448x448 pixel dimensions. We train the HRN model for 100 epochs with a batch size of 8 and an initial learning rate of 0.001. At test time, we use the trained YOLOv5l model for predicting the vehicle bounding boxes and then use the predictions to crop ROIs for HRN's input. 

We also experiment with two baseline detectors for the FGVD task. Firstly, we train the Faster-RCNN model on the FGVD's level-3 labels for 100 epochs with a batch size of 8 and an initial learning rate of 0.001. We use the Faster-RCNN with the resenet-50 backbone, and the model is pre-trained on the COCO detection dataset while taking an input image size of 512x512. Secondly, compared to the Faster-RCNN detector model, we train a similar sized YOLOv5-large variant model on level 3 labels for 100 epochs with a batch size of 16. 
We train the YOLOv5l model with the same hyperparameter configuration and the same pre-trained model as used in the first experiment explained above. 

We evaluate the performance of our models using the mean Average Precision (mAP) metric on all three levels. For the baselines, we derive the mAPs for all the levels from the combined label (sample combined labels are shown below the zoomed-out vehicle crops in Fig.~\ref{fig:teaser} a) on which the detectors are trained. We use GeForce GTX $3080$ Ti GPU for all our experiments. We present the results of the above experiments in the next section.

\begin{table}[ht]
\begin{center}
\caption{FGVD detection results (in \%): The combination of YOLO5l and HRN significantly improves the mAPs on all levels compared to existing detectors.}\vspace{-2mm}

\begin{tabular}{cccc}
\hline
\textbf{Model} & \textbf{L-1 mAP} & \textbf{L-2 mAP} & \textbf{L-3 mAP}  \\\hline
F-RCNN        & 54.43            & 41.46            & 31.92                                                                                                   \\\hline
YOLOv5l       & 61.70            & 42.40            & 32.75                                                                                                    \\\hline
\bf{YOLOv5L + HRN}  & {\bf 83.21}            & {\bf 59.02}            & {\bf 48.40}                                                          \\\hline
\end{tabular}
\vspace{-2.5mm}
\label{table:detection_results}
\end{center}
\end{table}

\begin{figure*}[t]
\includegraphics[width=\linewidth]{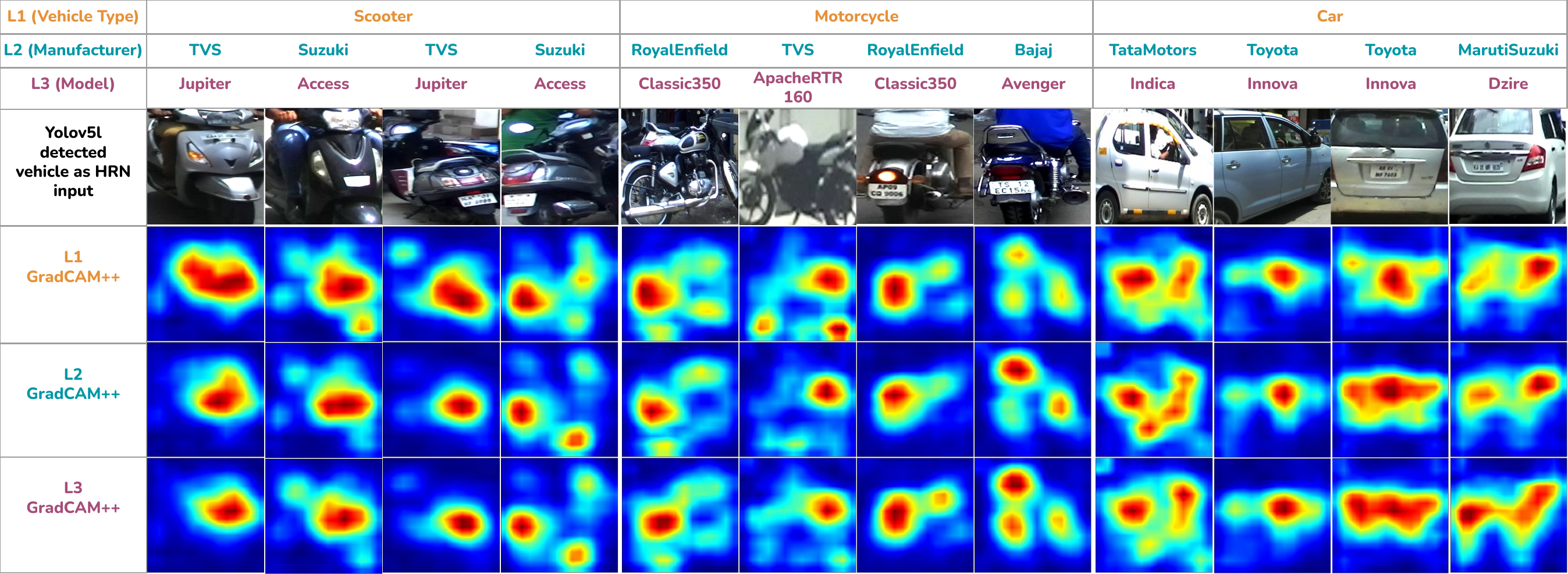}\Description{Diagram Figure}\vspace{-3mm}
\setlength{\belowcaptionskip}{-10pt}
\caption{Rows 1-3: correct predictions/labels obtained from YOLO+HRN with ROIs in row 4. GradCAM++~\cite{DBLP:journals/corr/abs-1710-11063} visualizations on L1, L2, and L3 levels (rows 5-8) from the HRN network, when applied on the row 4 ROIs. Multiple orientations of scooters, motorcycles, and cars are compared. From Row 4, Col 2: two front-view images of scooter models, followed by the rear-view images of the same models. Row 4, Col 6: two side-view motorcycle model images followed by two rear-view images. Row 4, Col 10: two side-view and two back-view images of different car models. The GradCAM++ visualizations highlight the distinguishing features for each ROI, like the front, side, and rear body structures, backlight/blinkers (most vehicles), silencer (Access rear-view and Classic 350), petrol tank (Jupiter rear-view), door handles, and engine shape for the corresponding vehicles.}
\label{fig:fgvd_with_gradcam}
\end{figure*}
\begin{figure}[t]
\includegraphics[width=\linewidth]{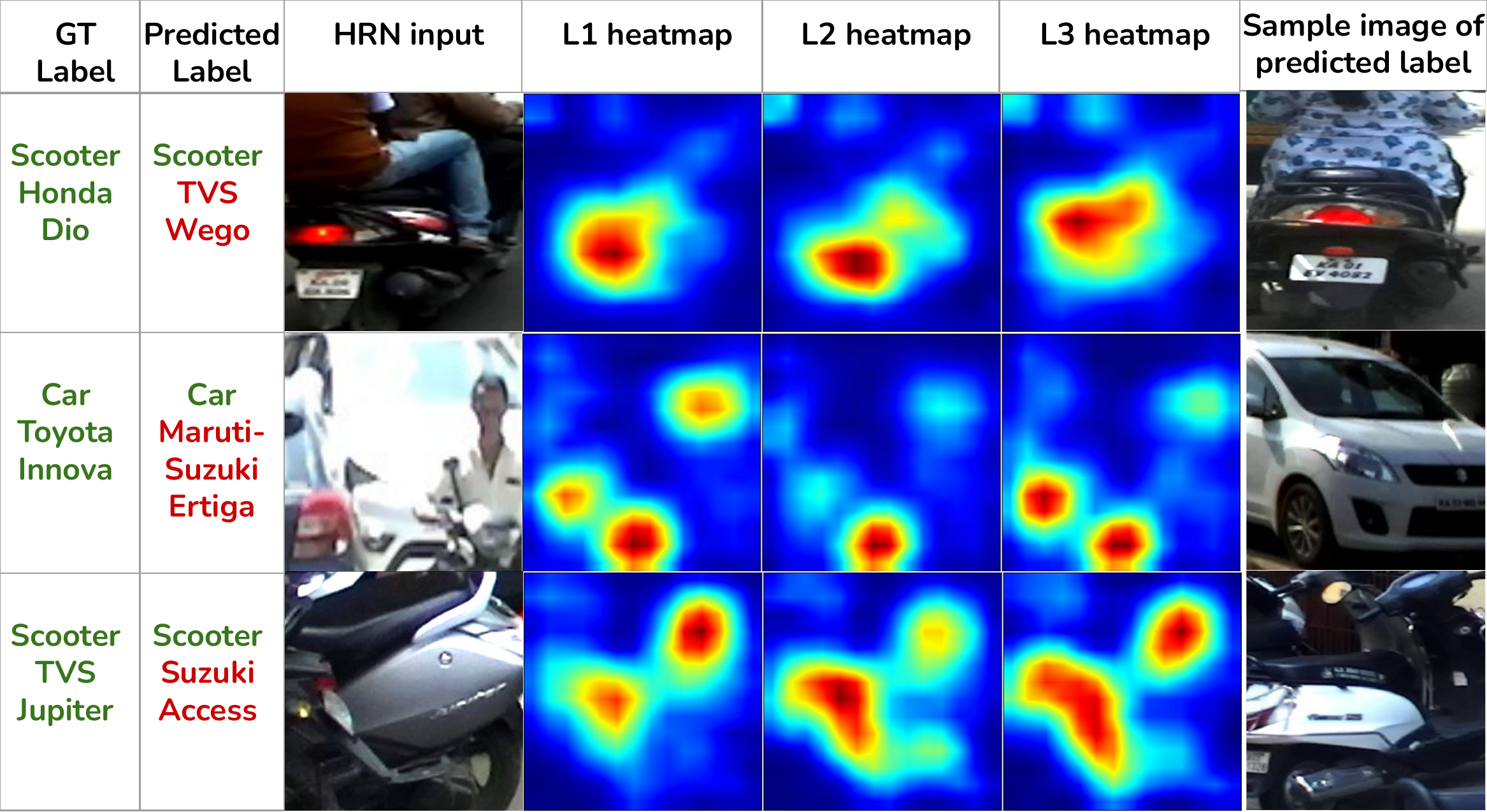}\Description{Diagram Figure}
\setlength{\belowcaptionskip}{-10pt}
\caption{Failure cases with their GradCAM++~\cite{DBLP:journals/corr/abs-1710-11063} visualizations for L1, L2, and L3 levels, along with images of an incorrectly predicted vehicle model for comparison (on the right) with the actual input image (on the left). Top and Bottom: Examples of high similarity in visual cues between the predicted and the actual classes. Mid: extreme illumination with occlusion from multiple objects. }
\label{fig:detectionfailures_with_gradcam}

\end{figure}

\section{RESULTS}\label{section:RESULTS} 
We show the results of our detection experiments in Table~\ref{table:detection_results}. As shown in its first row, with the Faster-RCNN model, we obtain the mAPs of $58.9\%$, $41.4\%$, and $31.9\%$ on level 1 (L-1), level 2 (L-2), and level 3 (L-3) labels, respectively. The Yolov5l model obtains mAPs of $61.7\%$, $42.4\%$, and $32.7\%$ on the three levels. We observe that the YOLOv5l detector performs significantly better than the Faster-RCNN model on all the hierarchical levels. We thus choose the YOLOv5l model with HRN for the final experiment. Using the YOLOv5l+HRN model, the mAP scores at L-1 and L-2 levels also improve by $21.5\%$ and $16.6\%$, respectively. We obtain an mAP of 48.4\% on the L-3 level, which is a substantial performance boost of $15.7\%$ compared to the standard YOLOv5l model.

We showcase the YOLOv5l+HRN model's detection results on sample images from the FGVD test set in Fig.~\ref{fig:fgvd_sample_outputs}. The classified fine-grained labels from the YOLOv5l+HRN model are shown alongside the predicted bounding boxes. The correctly classified predictions are shown in green, and the incorrect predictions are in red. The figure depicts our model's performance on images containing vehicles at multiple viewpoints and resolutions with occlusions and low visibility scenarios. It can be observed that the model is giving correct predictions for the occluded vehicles with low visibility, like the Hyundai Santro car in the left image of Fig.~\ref{fig:fgvd_sample_outputs}. Similarly, the Honda Shine motorcycle on the right image left side of the same figure is classified correctly at all three levels. We also showcase an occlusion scenario where the model gives an incorrect prediction. The right image in Fig.~\ref{fig:fgvd_sample_outputs}, shows that for the Hyundai Creta car (ground truth), our model gives incorrect prediction as Hyundai Santro car, which is highlighted in the red box. The zoomed-out views with adjusted brightness show the actual vehicle images clearly. For the incorrect detection in Fig.~\ref{fig:fgvd_sample_outputs}, we can observe that the model detects vehicle type and manufacturer correctly, even in the high occlusion scenario.

In Fig.~\ref{fig:fgvd_with_gradcam}, we showcase the GradCAM++~\cite{DBLP:journals/corr/abs-1710-11063} visualizations of the detected fine-grained vehicles from the YOLO-HRN model. On the figure's left side, we compare two front-view scooter images. The L1 GradCAM++ heatmap of the Jupiter model shows the model's focus on most of the front body parts along with its blinkers. The corresponding heatmap for the Access model also shows the focus on the front body part along with the mudguard region. While the L3 visualizations focus on very particular attributes for the Jupiter model, it is centered around the air vent part above the mudguard. For the Access model, the focus is on the shape of the front body part. Similarly, we compare the rear views of these vehicle models. The L1 heatmap of the Jupiter model shows a major focus on the rear body part with some highlights (in green) on its speedometer display. In contrast, for the Access sample, the primary focus is on the back body part with some highlights around the silencer, speedometer display, and the key region. Overall L1 heatmaps show the model's focus on most of the crucial body parts of the vehicle. While in the L3 visualizations of the Jupiter model the focus is centered around the petrol tank lid area just above the backlight, and for the Access, it is primarily centered around the backlight region with some highlights on the silencer part as well. These features highlighted by the L3 visualizations are the main feature points that help distinguish the two vehicle models. 

We have made similar comparisons between motorcycles and cars. We first compare the side views of the two models and then the rear views of the other two models. The L1 heatmap of the Classic350 model's side view shows the model's focus primarily around the suspension area just beside the triangular engine boundary, with some focus on the petrol tank, silencer, and the engine area at the bottom. While the L1 visualizations of the ApacheRTR160 TVS motorcycle focus on the vehicle's back body part, the back wheel's central area, and the front wheel. The L1 visualizations cover all the important key points that help identify a motorcycle, along with providing additional feature points for the classification at finer levels. The L2 visualizations highlight regions very similar to L3 visualizations, which ultimately help identify the correct L3 label. The L3 visualizations of the Classic350 model show the main focus on the suspension region and some on the silencer. For ApacheRTR160, the main focus is around the back body part. These focus points correspond to the unique feature points which are specific to that vehicle model only. Now we compare the rear views of Classic350 and Avenger models. These models' L1 and L3 heatmaps do not show much difference since only the back region of these vehicles are visible, where the number of critical key points is limited. The main distinguishing key point which is being highlighted in these heatmaps is around the backlight region and a little bit on the side body part, which is partially visible. 

Similarly, we compare the side views of cars. The L1 and L3 visualizations of the side views of the Indica and Innova models look nearly similar. For Indica, the door handle and the front window pillar regions are primarily focused, while the entire side body part is also highlighted in green. For Innova, the door handles and the window edges are primarily focused. Similarly, the rear views of Innova and Dzire show a primary focus on the shape and structure of the back body part along with the emphasis on the backlight region as visible in the Dzire model. Again due to the limited number of key points, we suspect that the highlighted parts in all the levels are similar, highlighting the entire overlapping regions considered for the L3 classification. Overall we showcase the hierarchical relationship between the coarse-level and fine-level labels and the overlap of similar features among them. Fig.~\ref{fig:detectionfailures_with_gradcam} shows the failure cases for similar vehicles in extreme illumination/occlusion scenarios.
\begin{table}[ht]
\begin{center}

\caption{Recognition accuracies (in \%) of HRN on standard fine-grained classification datasets and FGVD. For all three levels, the FGVD dataset remains the most challenging.}\vspace{-2mm}

\begin{tabular}{cccccc}
\hline
\textbf{Dataset} & \textbf{L-1} & \textbf{L-2} & \textbf{L-3} & \textbf{\# cls.} & \textbf{\# objs.}  \\\hline
Aircraft ~\cite{maji2013fine} &            97.45  &    95.79        & 92.58        & 100        & 10000                                                                                                 \\\hline
CUB-200-2011~\cite{wah2011caltech} &      98.67       &      95.51      & 86.60        & 200        & 11788                                                                                                 \\\hline
 Stanford Cars ~\cite{KrauseStarkDengFei-Fei_3DRR2013} &            97.41  &    94.03        & NA        & 207        & 16185                                                                                                 \\\hline
\bf{FGVD dataset} & {\bf 96.69}            & {\bf 79.44}            & {\bf 76.35}            & {\bf 217}            & {\bf 24450}                                                          \\\hline
\end{tabular}
\vspace{-2.5mm}
\label{table:classaccuraciesResults}
\end{center}
\end{table}

We also evaluate the L1, L2, and L3 label's classification accuracies using the fine-tuned HRN model on our test dataset and compare it with the corresponding state-of-the-art accuracies on standard fine-grained classification datasets as reported in the paper ~\cite{chen2022label}. We find that the Level-3 classification accuracy on our dataset is the lowest, i.e., 76.35\%, among other datasets, despite having more objects in FGVD, as shown in Table~\ref{table:classaccuraciesResults}. The reason for low accuracy on more fine-grained levels of FGVD is the high diversity in classes (217) and complexities like occlusions, non-object-centric images, pose variations, scale variations, lighting conditions, etc. The results quantitatively demonstrate the high complexity involved in the FGVD dataset and the challenges involved in the classification and detection of fine-grained vehicles in the wild. 

Although the YOLOv5l+HRN model performs better than other baseline detectors, there is still a massive gap of around $35\%$ mAP values between the coarse-level and the fine-grained level detection performances in Table~\ref{table:detection_results}. Thus, we propose this dataset for future research works related to the FGVD.

\section{Conclusion}
This paper presents the first dataset for the Fine-Grained Detection of Vehicles while also providing the baselines for the same. FGVD is also the first dataset in which the fine-grained labels for $5$ additional vehicle types are available apart from just cars. Our dataset can be used for vehicle re-identification in on-road surveillance systems, generating alarms for road safety systems, and promoting the development of ADAS products for Indian roads. Specifically, we showcase the uniqueness of our dataset regarding the detection of fine-grained vehicles in the wild compared to other related datasets. We also provide the results of the YOLOv5l+HRN model for the FGVD dataset, with which we obtain a $15.7\%$ gain in mAP over baseline detectors for the most complex level. For future works, we plan to fuse the architectures of detection and classification models to improve the overall efficiency of the solution.

{\bf Acknowledgement:} We thank iHubData, the Technology Innovation Hub (TIH) at IIIT-Hyderabad for supporting this project.





\bibliographystyle{ACM-Reference-Format}
\bibliography{ICVGIP21-CameraReady-Template}


\begin{thebibliography}{22}


\ifx \showCODEN    \undefined \def \showCODEN     #1{\unskip}     \fi
\ifx \showDOI      \undefined \def \showDOI       #1{#1}\fi
\ifx \showISBNx    \undefined \def \showISBNx     #1{\unskip}     \fi
\ifx \showISBNxiii \undefined \def \showISBNxiii  #1{\unskip}     \fi
\ifx \showISSN     \undefined \def \showISSN      #1{\unskip}     \fi
\ifx \showLCCN     \undefined \def \showLCCN      #1{\unskip}     \fi
\ifx \shownote     \undefined \def \shownote      #1{#1}          \fi
\ifx \showarticletitle \undefined \def \showarticletitle #1{#1}   \fi
\ifx \showURL      \undefined \def \showURL       {\relax}        \fi
\providecommand\bibfield[2]{#2}
\providecommand\bibinfo[2]{#2}
\providecommand\natexlab[1]{#1}
\providecommand\showeprint[2][]{arXiv:#2}

\bibitem[\protect\citeauthoryear{Chattopadhyay, Sarkar, Howlader, and
  Balasubramanian}{Chattopadhyay et~al\mbox{.}}{2017}]%
        {DBLP:journals/corr/abs-1710-11063}
\bibfield{author}{\bibinfo{person}{Aditya Chattopadhyay},
  \bibinfo{person}{Anirban Sarkar}, \bibinfo{person}{Prantik Howlader}, {and}
  \bibinfo{person}{Vineeth~N. Balasubramanian}.}
  \bibinfo{year}{2017}\natexlab{}.
\newblock \showarticletitle{Grad-CAM++: Generalized Gradient-based Visual
  Explanations for Deep Convolutional Networks}.
\newblock \bibinfo{journal}{\emph{CoRR}}  \bibinfo{volume}{abs/1710.11063}
  (\bibinfo{year}{2017}).
\newblock
\showeprint[arXiv]{1710.11063}
\urldef\tempurl%
\url{http://arxiv.org/abs/1710.11063}
\showURL{%
\tempurl}


\bibitem[\protect\citeauthoryear{Chen, Wang, Liu, and Qian}{Chen
  et~al\mbox{.}}{2022}]%
        {chen2022label}
\bibfield{author}{\bibinfo{person}{Jingzhou Chen}, \bibinfo{person}{Peng Wang},
  \bibinfo{person}{Jian Liu}, {and} \bibinfo{person}{Yuntao Qian}.}
  \bibinfo{year}{2022}\natexlab{}.
\newblock \showarticletitle{Label {Relation Graphs Enhanced Hierarchical
  Residual Network for Hierarchical Multi-Granularity C}lassification}. In
  \bibinfo{booktitle}{\emph{Proceedings of the IEEE/CVF Conference on Computer
  Vision and Pattern Recognition}}. \bibinfo{pages}{4858--4867}.
\newblock


\bibitem[\protect\citeauthoryear{Deng, Dong, Socher, Li, Li, and Fei-Fei}{Deng
  et~al\mbox{.}}{2009}]%
        {5206848}
\bibfield{author}{\bibinfo{person}{Jia Deng}, \bibinfo{person}{Wei Dong},
  \bibinfo{person}{Richard Socher}, \bibinfo{person}{Li-Jia Li},
  \bibinfo{person}{Kai Li}, {and} \bibinfo{person}{Li Fei-Fei}.}
  \bibinfo{year}{2009}\natexlab{}.
\newblock \showarticletitle{ImageNet: A large-scale hierarchical image
  database}. In \bibinfo{booktitle}{\emph{2009 IEEE Conference on Computer
  Vision and Pattern Recognition}}. \bibinfo{pages}{248--255}.
\newblock
\urldef\tempurl%
\url{https://doi.org/10.1109/CVPR.2009.5206848}
\showDOI{\tempurl}


\bibitem[\protect\citeauthoryear{He, Zhang, Ren, and Sun}{He
  et~al\mbox{.}}{2015}]%
        {DBLP:journals/corr/HeZRS15}
\bibfield{author}{\bibinfo{person}{Kaiming He}, \bibinfo{person}{Xiangyu
  Zhang}, \bibinfo{person}{Shaoqing Ren}, {and} \bibinfo{person}{Jian Sun}.}
  \bibinfo{year}{2015}\natexlab{}.
\newblock \showarticletitle{Deep Residual Learning for Image Recognition}.
\newblock \bibinfo{journal}{\emph{CoRR}}  \bibinfo{volume}{abs/1512.03385}
  (\bibinfo{year}{2015}).
\newblock
\showeprint[arXiv]{1512.03385}
\urldef\tempurl%
\url{http://arxiv.org/abs/1512.03385}
\showURL{%
\tempurl}


\bibitem[\protect\citeauthoryear{Jocher, Stoken, Borovec, NanoCode012,
  ChristopherSTAN, Changyu, Laughing, tkianai, Hogan, lorenzomammana, yxNONG,
  AlexWang1900, Diaconu, Marc, wanghaoyang0106, ml5ah, Doug, Ingham, Frederik,
  Guilhen, Hatovix, Poznanski, Fang, Yu, changyu98, Wang, Gupta, Akhtar,
  PetrDvoracek, and Rai}{Jocher et~al\mbox{.}}{2020}]%
        {yolov5}
\bibfield{author}{\bibinfo{person}{Glenn Jocher}, \bibinfo{person}{Alex
  Stoken}, \bibinfo{person}{Jirka Borovec}, \bibinfo{person}{NanoCode012},
  \bibinfo{person}{ChristopherSTAN}, \bibinfo{person}{Liu Changyu},
  \bibinfo{person}{Laughing}, \bibinfo{person}{tkianai}, \bibinfo{person}{Adam
  Hogan}, \bibinfo{person}{lorenzomammana}, \bibinfo{person}{yxNONG},
  \bibinfo{person}{AlexWang1900}, \bibinfo{person}{Laurentiu Diaconu},
  \bibinfo{person}{Marc}, \bibinfo{person}{wanghaoyang0106},
  \bibinfo{person}{ml5ah}, \bibinfo{person}{Doug}, \bibinfo{person}{Francisco
  Ingham}, \bibinfo{person}{Frederik}, \bibinfo{person}{Guilhen},
  \bibinfo{person}{Hatovix}, \bibinfo{person}{Jake Poznanski},
  \bibinfo{person}{Jiacong Fang}, \bibinfo{person}{Lijun Yu},
  \bibinfo{person}{changyu98}, \bibinfo{person}{Mingyu Wang},
  \bibinfo{person}{Naman Gupta}, \bibinfo{person}{Osama Akhtar},
  \bibinfo{person}{PetrDvoracek}, {and} \bibinfo{person}{Prashant Rai}.}
  \bibinfo{year}{2020}\natexlab{}.
\newblock \bibinfo{booktitle}{\emph{{ultralytics/yolov5: v3.1 - Bug Fixes and
  Performance Improvements}}}.
\newblock
\urldef\tempurl%
\url{https://doi.org/10.5281/zenodo.4154370}
\showDOI{\tempurl}


\bibitem[\protect\citeauthoryear{Khosla, Jayadevaprakash, Yao, and
  Fei-Fei}{Khosla et~al\mbox{.}}{2011}]%
        {KhoslaYaoJayadevaprakashFeiFei_FGVC2011}
\bibfield{author}{\bibinfo{person}{Aditya Khosla}, \bibinfo{person}{Nityananda
  Jayadevaprakash}, \bibinfo{person}{Bangpeng Yao}, {and} \bibinfo{person}{Li
  Fei-Fei}.} \bibinfo{year}{2011}\natexlab{}.
\newblock \showarticletitle{Novel Dataset for Fine-Grained Image
  Categorization}. In \bibinfo{booktitle}{\emph{First Workshop on Fine-Grained
  Visual Categorization, IEEE Conference on Computer Vision and Pattern
  Recognition}}. \bibinfo{address}{Colorado Springs, CO}.
\newblock


\bibitem[\protect\citeauthoryear{Krause, Stark, Deng, and Fei-Fei}{Krause
  et~al\mbox{.}}{2013}]%
        {KrauseStarkDengFei-Fei_3DRR2013}
\bibfield{author}{\bibinfo{person}{Jonathan Krause}, \bibinfo{person}{Michael
  Stark}, \bibinfo{person}{Jia Deng}, {and} \bibinfo{person}{Li Fei-Fei}.}
  \bibinfo{year}{2013}\natexlab{}.
\newblock \showarticletitle{3D Object Representations for Fine-Grained
  Categorization}. In \bibinfo{booktitle}{\emph{4th International IEEE Workshop
  on 3D Representation and Recognition (3dRR-13)}}. \bibinfo{address}{Sydney,
  Australia}.
\newblock


\bibitem[\protect\citeauthoryear{Lin, Maire, Belongie, Bourdev, Girshick, Hays,
  Perona, Ramanan, Doll{\'{a}}r, and Zitnick}{Lin et~al\mbox{.}}{2014}]%
        {DBLP:journals/corr/LinMBHPRDZ14}
\bibfield{author}{\bibinfo{person}{Tsung{-}Yi Lin}, \bibinfo{person}{Michael
  Maire}, \bibinfo{person}{Serge~J. Belongie}, \bibinfo{person}{Lubomir~D.
  Bourdev}, \bibinfo{person}{Ross~B. Girshick}, \bibinfo{person}{James Hays},
  \bibinfo{person}{Pietro Perona}, \bibinfo{person}{Deva Ramanan},
  \bibinfo{person}{Piotr Doll{\'{a}}r}, {and} \bibinfo{person}{C.~Lawrence
  Zitnick}.} \bibinfo{year}{2014}\natexlab{}.
\newblock \showarticletitle{Microsoft {COCO:} Common Objects in Context}.
\newblock \bibinfo{journal}{\emph{CoRR}}  \bibinfo{volume}{abs/1405.0312}
  (\bibinfo{year}{2014}).
\newblock
\showeprint[arXiv]{1405.0312}
\urldef\tempurl%
\url{http://arxiv.org/abs/1405.0312}
\showURL{%
\tempurl}


\bibitem[\protect\citeauthoryear{Maji, Rahtu, Kannala, Blaschko, and
  Vedaldi}{Maji et~al\mbox{.}}{2013}]%
        {maji2013fine}
\bibfield{author}{\bibinfo{person}{Subhransu Maji}, \bibinfo{person}{Esa
  Rahtu}, \bibinfo{person}{Juho Kannala}, \bibinfo{person}{Matthew Blaschko},
  {and} \bibinfo{person}{Andrea Vedaldi}.} \bibinfo{year}{2013}\natexlab{}.
\newblock \showarticletitle{Fine-grained visual classification of aircraft}.
\newblock \bibinfo{journal}{\emph{arXiv preprint arXiv:1306.5151}}
  (\bibinfo{year}{2013}).
\newblock


\bibitem[\protect\citeauthoryear{Najeeb, Raza, Yusuf, and Sultan}{Najeeb
  et~al\mbox{.}}{2022}]%
        {najeeb2022fine}
\bibfield{author}{\bibinfo{person}{Syeda~Aneeba Najeeb},
  \bibinfo{person}{Rana~Hammad Raza}, \bibinfo{person}{Adeel Yusuf}, {and}
  \bibinfo{person}{Zamra Sultan}.} \bibinfo{year}{2022}\natexlab{}.
\newblock \showarticletitle{Fine-grained vehicle classification in urban
  traffic scenes using deep learning}. In \bibinfo{booktitle}{\emph{Proceedings
  of the 11th International Conference on Robotics, Vision, Signal Processing
  and Power Applications}}. Springer, \bibinfo{pages}{902--908}.
\newblock


\bibitem[\protect\citeauthoryear{Nilsback and Zisserman}{Nilsback and
  Zisserman}{2008}]%
        {nilsback2008automated}
\bibfield{author}{\bibinfo{person}{Maria-Elena Nilsback} {and}
  \bibinfo{person}{Andrew Zisserman}.} \bibinfo{year}{2008}\natexlab{}.
\newblock \showarticletitle{Automated flower classification over a large number
  of classes}. In \bibinfo{booktitle}{\emph{2008 Sixth Indian Conference on
  Computer Vision, Graphics \& Image Processing}}. IEEE,
  \bibinfo{pages}{722--729}.
\newblock


\bibitem[\protect\citeauthoryear{Redmon and Farhadi}{Redmon and
  Farhadi}{2018a}]%
        {yolov3}
\bibfield{author}{\bibinfo{person}{Joseph Redmon} {and} \bibinfo{person}{Ali
  Farhadi}.} \bibinfo{year}{2018}\natexlab{a}.
\newblock \showarticletitle{YOLOv3: An Incremental Improvement}.
\newblock  (\bibinfo{date}{04} \bibinfo{year}{2018}).
\newblock


\bibitem[\protect\citeauthoryear{Redmon and Farhadi}{Redmon and
  Farhadi}{2018b}]%
        {DBLP:journals/corr/abs-1804-02767}
\bibfield{author}{\bibinfo{person}{Joseph Redmon} {and} \bibinfo{person}{Ali
  Farhadi}.} \bibinfo{year}{2018}\natexlab{b}.
\newblock \showarticletitle{YOLOv3: An Incremental Improvement}.
\newblock \bibinfo{journal}{\emph{CoRR}}  \bibinfo{volume}{abs/1804.02767}
  (\bibinfo{year}{2018}).
\newblock
\showeprint[arXiv]{1804.02767}
\urldef\tempurl%
\url{http://arxiv.org/abs/1804.02767}
\showURL{%
\tempurl}


\bibitem[\protect\citeauthoryear{Ren, He, Girshick, and Sun}{Ren
  et~al\mbox{.}}{2015}]%
        {fasterRCNN}
\bibfield{author}{\bibinfo{person}{Shaoqing Ren}, \bibinfo{person}{Kaiming He},
  \bibinfo{person}{Ross Girshick}, {and} \bibinfo{person}{Jian Sun}.}
  \bibinfo{year}{2015}\natexlab{}.
\newblock \showarticletitle{Faster R-CNN: Towards Real-Time Object Detection
  with Region Proposal Networks}. In \bibinfo{booktitle}{\emph{Proceedings of
  the 28th International Conference on Neural Information Processing Systems -
  Volume 1}} (Montreal, Canada) \emph{(\bibinfo{series}{NIPS'15})}.
  \bibinfo{publisher}{MIT Press}, \bibinfo{address}{Cambridge, MA, USA},
  \bibinfo{pages}{91–99}.
\newblock


\bibitem[\protect\citeauthoryear{Sochor, Špaňhel, and Herout}{Sochor
  et~al\mbox{.}}{2018}]%
        {Sochor2018}
\bibfield{author}{\bibinfo{person}{J. Sochor}, \bibinfo{person}{J. Špaňhel},
  {and} \bibinfo{person}{A. Herout}.} \bibinfo{year}{2018}\natexlab{}.
\newblock \showarticletitle{BoxCars: Improving Fine-Grained Recognition of
  Vehicles Using 3-D Bounding Boxes in Traffic Surveillance}.
\newblock \bibinfo{journal}{\emph{IEEE Transactions on Intelligent
  Transportation Systems}} \bibinfo{volume}{PP}, \bibinfo{number}{99}
  (\bibinfo{year}{2018}), \bibinfo{pages}{1--12}.
\newblock
\showISSN{1524-9050}
\urldef\tempurl%
\url{https://doi.org/10.1109/TITS.2018.2799228}
\showDOI{\tempurl}


\bibitem[\protect\citeauthoryear{Sun, Kretzschmar, Dotiwalla, Chouard, Patnaik,
  Tsui, Guo, Zhou, Chai, Caine, Vasudevan, Han, Ngiam, Zhao, Timofeev,
  Ettinger, Krivokon, Gao, Joshi, Zhang, Shlens, Chen, and Anguelov}{Sun
  et~al\mbox{.}}{2020}]%
        {Sun_2020_CVPR}
\bibfield{author}{\bibinfo{person}{Pei Sun}, \bibinfo{person}{Henrik
  Kretzschmar}, \bibinfo{person}{Xerxes Dotiwalla}, \bibinfo{person}{Aurelien
  Chouard}, \bibinfo{person}{Vijaysai Patnaik}, \bibinfo{person}{Paul Tsui},
  \bibinfo{person}{James Guo}, \bibinfo{person}{Yin Zhou},
  \bibinfo{person}{Yuning Chai}, \bibinfo{person}{Benjamin Caine},
  \bibinfo{person}{Vijay Vasudevan}, \bibinfo{person}{Wei Han},
  \bibinfo{person}{Jiquan Ngiam}, \bibinfo{person}{Hang Zhao},
  \bibinfo{person}{Aleksei Timofeev}, \bibinfo{person}{Scott Ettinger},
  \bibinfo{person}{Maxim Krivokon}, \bibinfo{person}{Amy Gao},
  \bibinfo{person}{Aditya Joshi}, \bibinfo{person}{Yu Zhang},
  \bibinfo{person}{Jonathon Shlens}, \bibinfo{person}{Zhifeng Chen}, {and}
  \bibinfo{person}{Dragomir Anguelov}.} \bibinfo{year}{2020}\natexlab{}.
\newblock \showarticletitle{Scalability in Perception for Autonomous Driving:
  Waymo Open Dataset}. In \bibinfo{booktitle}{\emph{Proceedings of the IEEE/CVF
  Conference on Computer Vision and Pattern Recognition (CVPR)}}.
\newblock


\bibitem[\protect\citeauthoryear{Van~Horn, Branson, Farrell, Haber, Barry,
  Ipeirotis, Perona, and Belongie}{Van~Horn et~al\mbox{.}}{2015}]%
        {Horn_2015_CVPR}
\bibfield{author}{\bibinfo{person}{Grant Van~Horn}, \bibinfo{person}{Steve
  Branson}, \bibinfo{person}{Ryan Farrell}, \bibinfo{person}{Scott Haber},
  \bibinfo{person}{Jessie Barry}, \bibinfo{person}{Panos Ipeirotis},
  \bibinfo{person}{Pietro Perona}, {and} \bibinfo{person}{Serge Belongie}.}
  \bibinfo{year}{2015}\natexlab{}.
\newblock \showarticletitle{Building a Bird Recognition App and Large Scale
  Dataset With Citizen Scientists: The Fine Print in Fine-Grained Dataset
  Collection}. In \bibinfo{booktitle}{\emph{Proceedings of the IEEE Conference
  on Computer Vision and Pattern Recognition (CVPR)}}.
\newblock


\bibitem[\protect\citeauthoryear{Varma, Subramanian, Namboodiri, Chandraker,
  and Jawahar}{Varma et~al\mbox{.}}{2019}]%
        {varma2019idd}
\bibfield{author}{\bibinfo{person}{Girish Varma}, \bibinfo{person}{Anbumani
  Subramanian}, \bibinfo{person}{Anoop Namboodiri}, \bibinfo{person}{Manmohan
  Chandraker}, {and} \bibinfo{person}{CV Jawahar}.}
  \bibinfo{year}{2019}\natexlab{}.
\newblock \showarticletitle{I{DD: A Dataset for Exploring Problems of
  Autonomous Navigation in Unconstrained E}nvironments}. In
  \bibinfo{booktitle}{\emph{2019 IEEE Winter Conference on Applications of
  Computer Vision (WACV)}}. IEEE, \bibinfo{pages}{1743--1751}.
\newblock


\bibitem[\protect\citeauthoryear{Wah, Branson, Welinder, Perona, and
  Belongie}{Wah et~al\mbox{.}}{2011}]%
        {wah2011caltech}
\bibfield{author}{\bibinfo{person}{Catherine Wah}, \bibinfo{person}{Steve
  Branson}, \bibinfo{person}{Peter Welinder}, \bibinfo{person}{Pietro Perona},
  {and} \bibinfo{person}{Serge Belongie}.} \bibinfo{year}{2011}\natexlab{}.
\newblock \showarticletitle{The caltech-ucsd birds-200-2011 dataset}.
\newblock  (\bibinfo{year}{2011}).
\newblock


\bibitem[\protect\citeauthoryear{Wang, Liao, Yeh, Wu, Chen, and Hsieh}{Wang
  et~al\mbox{.}}{2019}]%
        {DBLP:journals/corr/abs-1911-11929}
\bibfield{author}{\bibinfo{person}{Chien{-}Yao Wang},
  \bibinfo{person}{Hong{-}Yuan~Mark Liao}, \bibinfo{person}{I{-}Hau Yeh},
  \bibinfo{person}{Yueh{-}Hua Wu}, \bibinfo{person}{Ping{-}Yang Chen}, {and}
  \bibinfo{person}{Jun{-}Wei Hsieh}.} \bibinfo{year}{2019}\natexlab{}.
\newblock \showarticletitle{CSPNet: {A} New Backbone that can Enhance Learning
  Capability of {CNN}}.
\newblock \bibinfo{journal}{\emph{CoRR}}  \bibinfo{volume}{abs/1911.11929}
  (\bibinfo{year}{2019}).
\newblock
\showeprint[arXiv]{1911.11929}
\urldef\tempurl%
\url{http://arxiv.org/abs/1911.11929}
\showURL{%
\tempurl}


\bibitem[\protect\citeauthoryear{Yang, Luo, Change~Loy, and Tang}{Yang
  et~al\mbox{.}}{2015}]%
        {Yang_2015_CVPR}
\bibfield{author}{\bibinfo{person}{Linjie Yang}, \bibinfo{person}{Ping Luo},
  \bibinfo{person}{Chen Change~Loy}, {and} \bibinfo{person}{Xiaoou Tang}.}
  \bibinfo{year}{2015}\natexlab{}.
\newblock \showarticletitle{A Large-Scale Car Dataset for Fine-Grained
  Categorization and Verification}. In \bibinfo{booktitle}{\emph{Proceedings of
  the IEEE Conference on Computer Vision and Pattern Recognition (CVPR)}}.
\newblock


\bibitem[\protect\citeauthoryear{Yu, Chen, Wang, Xian, Chen, Liu, Madhavan, and
  Darrell}{Yu et~al\mbox{.}}{2020}]%
        {yu2020bdd100k}
\bibfield{author}{\bibinfo{person}{Fisher Yu}, \bibinfo{person}{Haofeng Chen},
  \bibinfo{person}{Xin Wang}, \bibinfo{person}{Wenqi Xian},
  \bibinfo{person}{Yingying Chen}, \bibinfo{person}{Fangchen Liu},
  \bibinfo{person}{Vashisht Madhavan}, {and} \bibinfo{person}{Trevor Darrell}.}
  \bibinfo{year}{2020}\natexlab{}.
\newblock \showarticletitle{Bdd100k: A diverse driving dataset for
  heterogeneous multitask learning}. In \bibinfo{booktitle}{\emph{Proceedings
  of the IEEE/CVF conference on computer vision and pattern recognition}}.
  \bibinfo{pages}{2636--2645}.
\newblock


\end{thebibliography}

\end{document}